\pdfoutput=1

\documentclass[11pt]{article}
\usepackage{acl}
\usepackage{times}
\usepackage{latexsym}

\usepackage[T1]{fontenc}
\usepackage[utf8]{inputenc}
\usepackage{microtype}
\usepackage{inconsolata}
\usepackage[linesnumbered,ruled,vlined]{algorithm2e} 
\usepackage{setspace} 

\usepackage{amsthm}

\usepackage{amsmath}
\usepackage{graphicx}
\usepackage[noend]{algpseudocode}

\usepackage{enumitem}
\usepackage[linesnumbered,ruled,vlined]{algorithm2e}
\usepackage{multirow}
\usepackage{adjustbox}
\usepackage{bbm}
\usepackage{wrapfig,lipsum}
\usepackage{xspace}
\usepackage{booktabs,cellspace}  
\usepackage[normalem]{ulem}
\usepackage{makecell}
\usepackage{amssymb}
\usepackage{pifont}
\usepackage{todonotes}
\usepackage{lipsum}
\usepackage{afterpage}
\usepackage{float} 

\newcommand\blankpage{%
    \null
    \thispagestyle{empty}%
    \addtocounter{page}{-1}%
    \newpage}

\usepackage{microtype}

\usepackage{cleveref}

\newcommand{\gpt}{\textsc{GPT-3.5-Turbo}\xspace}

\newcommand\blfootnote[1]{%
  \begingroup
  \renewcommand\thefootnote{}\footnote{#1}%
  \addtocounter{footnote}{-1}%
  \endgroup
}

\newcommand{\up}[1]{\textcolor{red}{\ensuremath{\scriptstyle\uparrow#1}}}
\newcommand{\down}[1]{\textcolor{blue}{\ensuremath{\scriptstyle\downarrow#1}}}

\usepackage{etoolbox}
\usepackage[tikz]{bclogo}

\usepackage{subcaption}
\usepackage{adjustbox}

\definecolor{msftBlue}{RGB}{0,164,239}
\definecolor{msftGreen}{RGB}{127,186,0}
\definecolor{msftYello}{RGB}{255,185,0}
\definecolor{msftBlack}{RGB}{0,0,0}

\usepackage{tcolorbox} 
\tcbuselibrary{skins} 
\usepackage[T1]{fontenc}

\usepackage{tcolorbox} 
\tcbuselibrary{skins} 
\usepackage[T1]{fontenc}

\tcbset{
    userstyle/.style={
        enhanced,
        colback=white,
        colframe=black,
        colbacktitle=gray!20,
        coltitle=black,
        rounded corners,
        sharp corners=north,
        boxrule=0.5pt,
        drop shadow=black!50!white,
        attach boxed title to top left={
            xshift=-2mm,
            yshift=-2mm
        },
        boxed title style={
            rounded corners,
            size=small,
            colback=gray!20
        }
    },
    replystyleg/.style={
        enhanced,
        colback=green!15,
        colframe=black,
        colbacktitle=green!30,
        coltitle=black,
        boxrule=0.5pt,
        drop shadow=black!50!white,
        rounded corners,
        sharp corners=north,
        attach boxed title to top right={
            xshift=-2mm,
            yshift=-2mm
        },
        boxed title style={
            rounded corners,
            size=small,
            colback=green!40
        }
    },
    replystyler/.style={
        enhanced,
        colback=red!15,
        colframe=black,
        colbacktitle=red!40,
        coltitle=black,
        boxrule=0.5pt,
        drop shadow=black!50!white,
        rounded corners,
        sharp corners=north,
        attach boxed title to top right={
            xshift=-2mm,
            yshift=-2mm
        },
        boxed title style={
            rounded corners,
            size=small,
            colback=red!40
        }
    }
}
\newtcolorbox{userquery}[1][]{
    userstyle,
    title=Probed Fuses for \gpt,
    #1
}

\title{DiMo-GUI: Advancing Test-time Scaling in GUI Grounding via Modality-Aware Visual Reasoning}

\author{
\bf Hang Wu$^{1,3}$, Hongkai Chen$^{3\dagger}$, Yujun Cai$^2$, Chang Liu$^{3}$, \\
\bf Qingwen Ye$^3$, Ming-Hsuan Yang$^1$, Yiwei Wang$^1$ \\
$^1$University of California, Merced,
$^2$The University of Queensland,  \\
$^3$vivo Mobile Communication Co., Ltd \\
\texttt{hangwu@ucmerced.edu, allenhkchen@gmail.com} \\
\href{https://wuhang03.github.io/DiMo-GUI-homepage/}{\textcolor{magenta}{\texttt{https://wuhang03.github.io/DiMo-GUI-homepage/}}}
\url{}
}
\date{}

\begin{document}
\maketitle

\blfootnote{$^\dagger$The corresponding author.}

\begin{abstract}
Grounding natural language queries in graphical user interfaces (GUIs) poses unique challenges due to the diversity of visual elements, spatial clutter, and the ambiguity of language. In this paper, we introduce DiMo-GUI, a training-free framework for GUI grounding that leverages two core strategies: dynamic visual grounding and modality-aware optimization. Instead of treating the GUI as a monolithic image, our method splits the input into textual elements and iconic elements, allowing the model to reason over each modality independently using general-purpose vision-language models. When predictions are ambiguous or incorrect, DiMo-GUI dynamically focuses attention by generating candidate focal regions centered on the model’s initial predictions and incrementally zooms into subregions to refine the grounding result. This hierarchical refinement process helps disambiguate visually crowded layouts without the need for additional training or annotations. We evaluate our approach on standard GUI grounding benchmarks and demonstrate consistent improvements over baseline inference pipelines, highlighting the effectiveness of combining modality separation with region-focused reasoning. 

\end{abstract}


\begin{figure}
    \centering
    \includegraphics[width=0.9\linewidth]{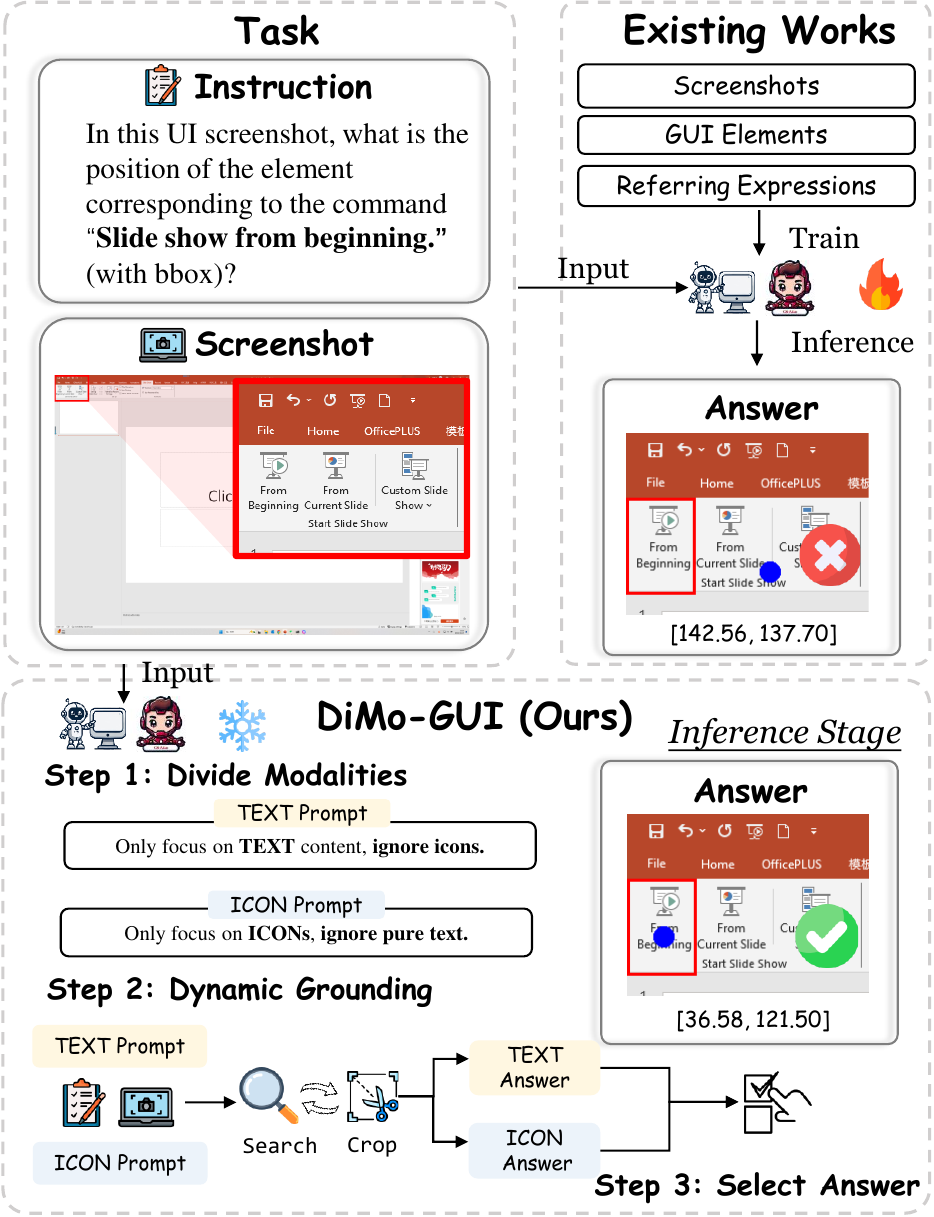}
    \vspace{-8pt}
    \caption{\textbf{Overview.} DiMo-GUI searches separately within the text and icon elements based on the instruction and the screenshot.}
    \label{fig:teaser}
    \vspace{-25pt}
\end{figure}

\section{Introduction}

Graphical user interface (GUI) agents play an increasingly central role in modern computing, allowing a wide range of applications, from automated web navigation to intuitive control of operating systems~\cite{anderson2018vision,liu2024visualagentbench}. With the rise of large-scale vision-language models (VLMs), recent research has focused on leveraging both visual and textual modalities to build more intelligent and interactive agents. However, many existing frameworks rely predominantly on text-based reasoning~\cite{yang2024agentoccam} or adopt simplistic visual grounding strategies~\cite{lu2024omniparser, gou2024navigation}. In practice, real-world GUIs often contain a large number of irrelevant or distracting elements—such as menu bars, advertisements, or extraneous buttons—that can overwhelm purely text-driven or naive visual approaches. This discrepancy between text-heavy inference and the complex visual nature of GUIs frequently results in errors, such as clicking incorrect buttons or navigating to unintended regions. Given that these agents are often tasked with high-level decision making, such low-level mistakes can accumulate, ultimately degrading overall performance and task success rates.

Recent work on GUI agents generally falls into two major paradigms: one centered on text-based reasoning and planning, and the other grounded in visual understanding through VLMs. Text-focused methods typically generate textual descriptions or bounding boxes for each visual element to inform action decisions~\cite{lu2024omniparser}. However, these approaches struggle with visually complex scenarios where text descriptions are ambiguous, incomplete, or fail to capture crucial visual cues—such as floating windows or dynamic pop-ups—even when assisted by accessibility trees. In contrast, vision-based pipelines~\cite{gou2024navigation, qin2025uitars} rely heavily on the grounding capabilities of VLMs, but are prone to errors such as clicking on empty or incorrect regions due to limitations in one-shot visual inference. Critically, these systems often lack an error correction mechanism; once a mistake occurs, it goes unaddressed, compounding over time and leading to cascading failures during multi-step interaction tasks.

To address these limitations, we propose DiMo-GUI, a training-free GUI grounding framework that integrates progressive zoom-in refinement and modality-specific processing. Instead of relying on a single forward pass, DiMo-GUI starts from a coarse prediction of the focal region and iteratively narrows the focus by refining bounding boxes around the target. Meanwhile, it separates textual and graphical components within the GUI and processes them with tailored strategies, allowing the agent to better handle diverse content types. As shown in Fig.~\ref{fig:teaser}, this design avoids the need for additional training, and can be plugged into existing GUI agents. Empirically, we find that this step-wise, disentangled grounding pipeline significantly improves robustness in visually cluttered or ambiguous environments, while maintaining compatibility with general-purpose VLMs.

\begin{figure}
    \centering
    \includegraphics[width=1.0\linewidth]{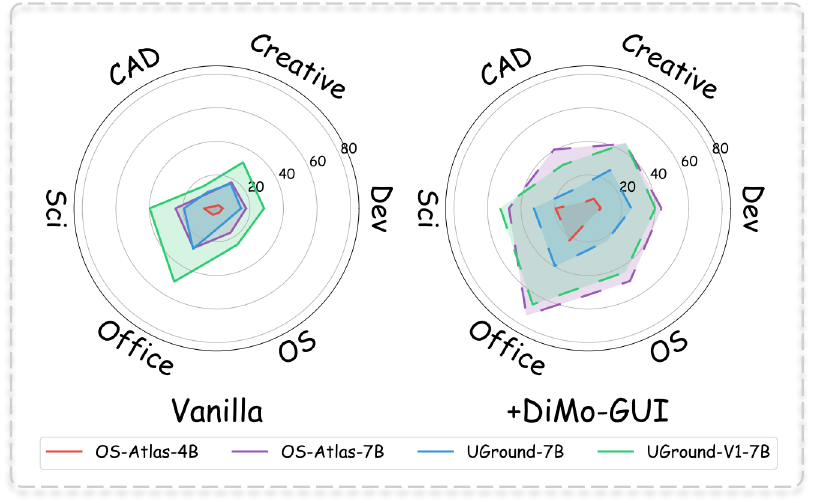}
    \vspace{-15pt} 
    \caption{\textbf{Grounding Performance of DiMo-GUI.} By integrating DiMo-GUI, existing models can achieve significant performance improvements on the dataset.}
    \vspace{-10pt}
    \label{fig:radar}
\end{figure}

We evaluate the proposed DiMo-GUI framework on the recently released ScreenSpot-Pro dataset by integrating it into several state-of-the-art models as reported in the original paper. Without modifying the model architecture or requiring any additional training, DiMo-GUI brings a clear performance improvement across key evaluation metrics as shown in Fig.~\ref{fig:radar}. These results demonstrate the effectiveness and generalizability of our training-free design in enhancing GUI grounding performance in existing large-scale models. Our main contributions can be summarized as follows:
\begin{itemize}
    \item We propose DiMo-GUI, a training-free framework that can be seamlessly integrated as a plug-and-play component into any GUI agent. Without requiring additional training or external data, DiMo-GUI effectively enhances grounding performance across various GUI tasks.
    \item DiMo-GUI introduces three key innovations: (1) a divide-and-conquer strategy that separates text and icon components for targeted processing, (2) a progressive zoom-in mechanism to increasingly focus on the target region, and (3) a dynamic halting system that enables timely decision-making and early  DiMo-GUIping to reduce overthinking and unnecessary computational cost.
    \item Extensive and comprehensive experiments demonstrate that DiMo-GUI can significantly enhance the grounding performance of various GUI agents across multiple benchmarks with minimal computational overhead, showcasing the effectiveness and generalizability of the proposed framework. 
\end{itemize}

\section{Related Work}

\subsection{GUI Agents}
Recent years have witnessed significant advances in GUI automation driven by large language models (LLMs). Early GUI agents predominantly focused on web interactions \cite{nakano2022webgpt, hong2024cogagent} and have gradually expanded to mobile \cite{zhang2023appagent, wang2024mobileagent} and desktop environments \cite{zhang2024ufo}. A fundamental challenge across these applications is precise element localization. Traditional approaches relied on structured information like XML and DOM trees \cite{zhang2023appagent}, but faced limitations in accessibility and information redundancy. Alternative methods using OCR \cite{du2020ppocr} or detection models \cite{liu2024groundingdino} introduced additional computational overhead. Recent advances in multimodal large language models (MLLMs) have enabled direct GUI element localization \cite{hong2024cogagent, cheng2024seeclick, lin2024showui}, partially bridging the visual perception gap. \cite{tang2025think} introduces a dual-system framework that combines fast prediction with systematic analysis to provide robust GUI foundation. OS-Atlas~\cite{wu2024osatlas} and UGround~\cite{gou2024navigation} created large datasets and trained models to handle out-of-distribution tasks. ~\cite{zhou2025gui, lee2025reguide, yuan2025enhancing, xia2025gui} explored improving grounding performance using reinforcement learning. ~\cite{tao2025understanding} proposes a framework and method to diagnose and reduce localization errors in MLLMs for GUI interaction, improving interpretability and robustness.

\subsection{Test-time scaling} 
Test-time scaling dynamically adjusts computational resources during inference to enhance model performance, with recent studies showing it can outperform increased train-time computation through strategies like best-of-N sampling and external verification~\cite{snell2024scaling, lee2025revise, hosseini2024vst}. In localization tasks, test-time scaling has also been framed as a search problem~\cite{wu2024vstar}. Inspired by its success in LLMs, similar techniques have been applied to GUI agents, such as leveraging action histories~\cite{zhang2023you}, gathering external information~\cite{nakano2022webgpt}, zooming in and searching~\cite{nguyen2024improved}, and adaptively refining focus regions~\cite{luo2025visual}. ~\cite{ge2025mrfd} proposes Multi-Region Fusion Decoding (MRFD), a training-free method that reduces hallucinations in LVLMs by leveraging inter-region consistency to improve factual grounding.

\begin{figure*}
    \centering
    \includegraphics[width=1.0\linewidth]{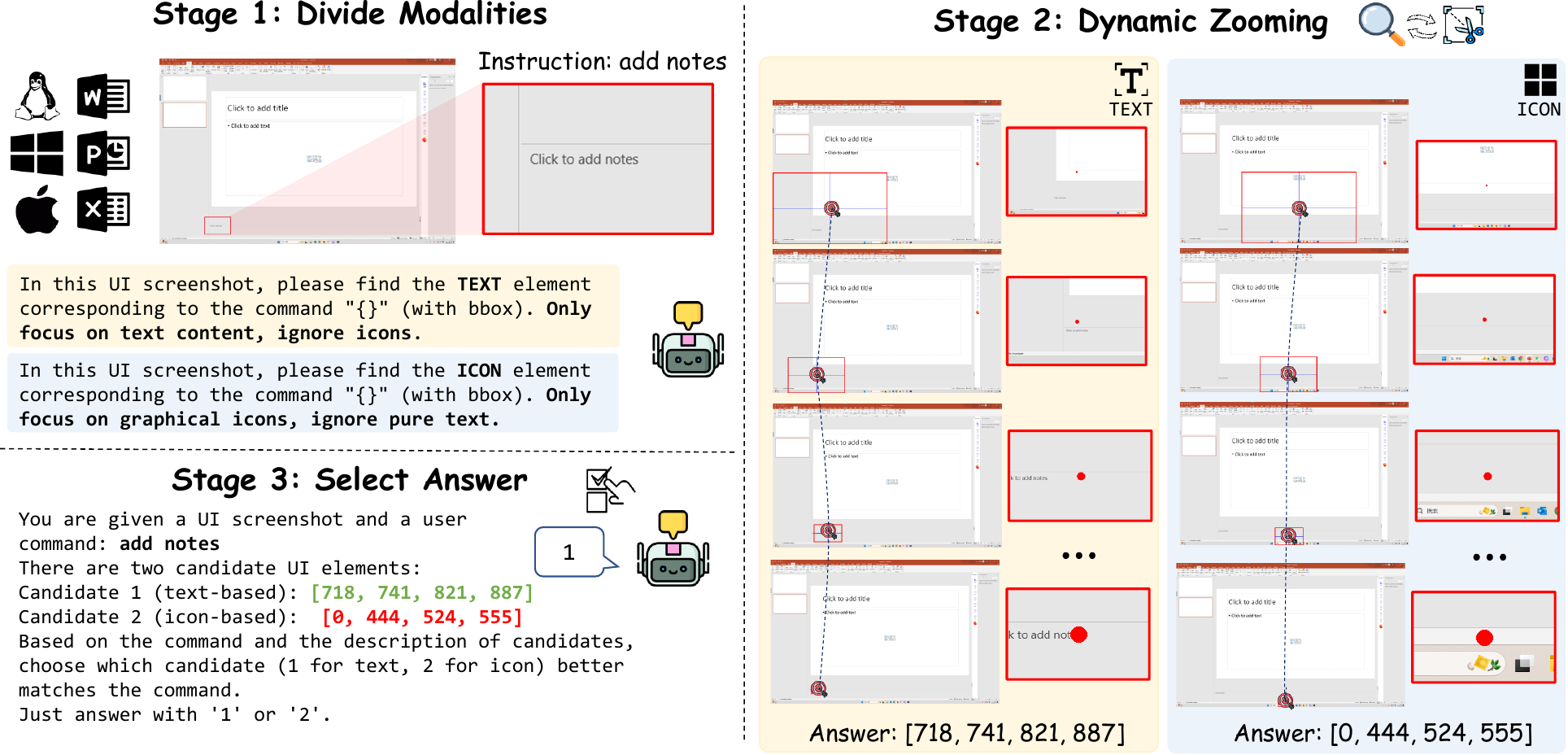}
    \caption{\textbf{Processing pipeline of DiMo-GUI.} DiMo-GUI decomposes the grounding process into three steps:
(1) Divide Modalities: It processes textual and icon elements in the screenshot separately to prevent interference between the two modalities.
(2) Dynamic Zooming: Based on an initial prediction, the model centers on the returned coordinates and crops a region half the size of the original image for more precise localization.
(3) Decision Making: By analyzing the instruction along with the screenshot, the model determines whether the text-based or icon-based candidate is more likely to be the correct answer.}
    \vspace{-10pt}
    \label{fig:enter-label}
\end{figure*}

\begin{algorithm}[t]
\begin{spacing}{1}
\caption{Dual-Modality Grounding with Dynamic Zooming}
\label{algorithm:DualModalityGrounding}
\LinesNumbered
\KwIn{Full-resolution GUI image $I$, instruction $Q$}
\KwOut{Final grounded coordinate $C^*$}

\textbf{Step 1: Text modality grounding.} \\
$C_{\text{text}} \leftarrow \texttt{DynamicGrounding}(I, Q, \text{``text''})$ \\

\textbf{Step 2: Icon modality grounding.} \\
$C_{\text{icon}} \leftarrow \texttt{DynamicGrounding}(I, Q, \text{``icon''})$ \\

\textbf{Step 3: Candidate selection.} \\
$C^* \leftarrow \texttt{Select}(C_{\text{text}}, C_{\text{icon}}, I, Q)$ \\
\Return $C^*$

\vspace{0.5em}
\textbf{Function:} \texttt{DynamicGrounding}($I$, $Q$, modality $m$)

Initialize zoom region: $R \leftarrow I$ \\
\For{$t = 1$ \KwTo max\_iters}{
    Predict coordinate:  \\
    $C_t \leftarrow \texttt{PredictCoordinate}(R, Q, m)$ \\
    \If{\texttt{ StopCondition}($C_t$, $t$)}{
        \Return $C_t$
    }
    $R \leftarrow \texttt{CropAround}(R, C_t)$ // update region
}
\Return $C_{\text{max}}$
\end{spacing}

\end{algorithm}

\section{Methodology}
To address the limitations of existing GUI agents in handling high-resolution images and their imbalanced performance between text and icon understanding, we propose a novel framework called DiMo-GUI. As shown in the algorithm~\ref{algorithm:DualModalityGrounding}, our method integrates a dynamic zooming mechanism and a modality decoupling strategy. Specifically, DiMo-GUI dynamically narrows down the target region through iterative zooming on the input high-resolution screenshot, progressively refining the localization until the target coordinates are identified. In parallel, DiMo-GUI decouples text-based and icon-based GUI elements, processing each modality independently to reduce cross-modal interference. This design mitigates a common shortcoming of vision-language models (VLMs), which typically exhibit stronger capabilities in text understanding compared to visual icon interpretation.

\subsection{Dynamic Grounding Mechanism}
High resolution remains one of the most significant challenges in GUI grounding, often leading to long inference times and excessive visual redundancy. A natural solution to this problem is to iteratively narrow down the target region, progressively refining the prediction of the target coordinates. To this end, DiMo-GUI introduces a dynamic zooming mechanism that enables efficient and focused localization. Specifically, the original high-resolution image is first passed to the model for an initial prediction. Based on the returned coordinates, a bounding box is cropped using the center point and a scaling factor of half the original image size. This cropped region is then used as input for the next round of inference.  Iterative zooming in allows the model to capture finer details of the target element, making it easier to recognize. At the same time, it significantly reduces redundant regions in the image, thereby increasing the signal-to-noise ratio. This helps the model receive less visual interference and focus more effectively on identifying the target element. As the iterations proceed, the model's attention becomes increasingly concentrated, ultimately enabling accurate target localization with minimal computational overhead.

The number of iterations in the zooming process plays a critical role in determining the final grounding performance. Since different GUI screenshots and user instructions vary in complexity, it is evident that a fixed number of iterations is not optimal for all cases. To address this, we introduce a dynamic iteration mechanism that allows the model to autonomously decide whether to  DiMo-GUI early during the progressive narrowing process. This approach not only reduces unnecessary iterations and improves inference efficiency but also prevents the model from "overthinking"—i.e., drifting into incorrect regions after having already located the correct target. Specifically, the method determines whether to continue zooming based on the spatial distance between the inference results before and after zooming. If the spatial distance between the predicted coordinates is smaller than one-sixth of the diagonal length of the pre-zoom image, it indicates that the target region has been localized with sufficient precision. In this case, further zooming is  stopped, and the final coordinates are returned as the result. The above process is described as \texttt{ StopCondition}($C_t$, $t$) in the algorithm~\ref{algorithm:DualModalityGrounding}, which decides whether to  stop dynamic zooming in the \textit{t} iteration based on the predicted coordinates $C_t$. Additionally, to prevent excessive zooming, we set an upper limit \textit{max\_iters} of seven zooming iterations.

\begin{table*}[!ht]
\setlength{\tabcolsep}{3.5pt} %
\centering
\resizebox{\linewidth}{!}{
\begin{tabular}{l|ccc|ccc|ccc|ccc|ccc|ccc|ccc}
\toprule
\textbf{Grounding Model} & \multicolumn{3}{c|}{\textbf{Development}} & \multicolumn{3}{c|}{\textbf{Creative}} & \multicolumn{3}{c|}{\textbf{CAD}} & \multicolumn{3}{c|}{\textbf{Scientific}} & \multicolumn{3}{c|}{\textbf{Office}} & \multicolumn{3}{c|}{\textbf{OS}} & \multicolumn{3}{c}{\textbf{Avg}} \\
\cmidrule(lr){2-4} \cmidrule(lr){5-7} \cmidrule(lr){8-10} \cmidrule(lr){11-13} \cmidrule(lr){14-16} \cmidrule(lr){17-19} \cmidrule(lr){20-22} 
 & \textbf{text} & \textbf{icon} & \textbf{avg} & \textbf{text} & \textbf{icon} & \textbf{avg} & \textbf{text} & \textbf{icon} & \textbf{avg} & \textbf{text} & \textbf{icon} & \textbf{avg} & \textbf{text} & \textbf{icon} & \textbf{avg} & \textbf{text} & \textbf{icon} & \textbf{avg} & \textbf{text} & \textbf{icon} & \textbf{avg} \\
\midrule
QwenVL-7B~\cite{bai2023qwenvl} & 0.0 & 0.0 & 0.0 & 0.0 & 0.0 & 0.0 & 0.0 & 0.0 & 0.0 & 0.7 & 0.0 & 0.4 & 0.0 & 0.0 & 0.0 & 0.0 & 0.0 & 0.0 & 0.1 & 0.0 & 0.1 \\
GPT-4o~\cite{openai2023gpt4} & 1.3 & 0.0 & 0.7 & 1.0 & 0.0 & 0.6 & 2.0 & 0.0 & 1.5 & 2.1 & 0.0 & 1.2 & 1.1 & 0.0 & 0.6 & 0.0 & 0.0 & 0.0 & 1.3 & 0.0 & 0.8 \\
SeeClick~\cite{cheng2024seeclick} & 0.6 & 0.0 & 0.3 & 1.0 & 0.0 & 0.6 & 2.5 & 0.0 & 1.9 & 3.5 & 0.0 & 2.0 & 1.1 & 0.0 & 0.5 & 2.8 & 0.0 & 1.5 & 1.8 & 0.0 & 1.1 \\
Qwen2-VL-7B~\cite{wang2024qwen2vl} & 2.6 & 0.0 & 1.3 & 1.5 & 0.0 & 0.9 & 0.5 & 0.0 & 0.4 & 6.3 & 0.0 & 3.5 & 3.4 & 1.9 & 3.0 & 0.9 & 0.0 & 0.5 & 2.5 & 0.2 & 1.6 \\
ShowUI-2B~\cite{lin2024showui} & 16.9 & 1.4 & 9.4 & 9.1 & 0.0 & 5.3 & 2.5 & 0.0 & 1.9 & 13.2 & 7.3 & 10.6 & 15.3 & 7.5 & 13.5 & 10.3 & 2.2 & 6.6 & 10.8 & 2.6 & 7.7 \\
CogAgent-18B~\cite{hong2024cogagent} & 14.9 & 0.7 & 8.0 & 9.6 & 0.0 & 5.6 & 7.1 & 3.1 & 6.1 & 22.2 & 1.8 & 13.4 & 13.0 & 0.0 & 6.5 & 5.6 & 0.0 & 3.1 & 12.0 & 0.8 & 7.7 \\
Aria-UI~\cite{yang2024ariaui} & 16.2 & 0.0 & 8.4 & 23.7 & 2.1 & 14.7 & 7.6 & 1.6 & 6.1 & 27.1 & 6.4 & 18.1 & 20.3 & 1.9 & 16.1 & 4.7 & 0.0 & 2.6 & 17.1 & 2.0 & 11.3 \\
Claude Comp.Use~\cite{hu2024claude} & 22.0 & 3.9 & 12.6 & 25.9 & 3.4 & 16.8 & 14.5 & 3.7 & 11.9 & 33.9 & 15.8 & 25.8 & 30.1 & 16.3 & 26.2 & 11.0 & 4.5 & 8.1 & 23.4 & 7.1 & 17.1 \\
UI-TARS-7B~\cite{qin2025uitars} & 58.4 & 12.4 & 36.1 & 50.0 & 9.1 & 32.8 & 20.8 & 9.4 & 18.0 & 63.9 & 31.8 & 50.0 & 63.3 & 20.8 & 53.5 & 30.8 & 16.9 & 24.5 & 47.8 & 16.2 & 35.7\\
UI-TARS-72B\cite{qin2025uitars} & 63.0 & 17.3 & 40.8 & 57.1 & 15.4 & 39.6 & 18.8 & 12.5 & 17.2 & 64.6 & 20.9 & 45.7 & 63.3 & 26.4 & 54.8 & 42.1 & 15.7 & 30.1 & 50.9 & 17.5 & 38.1 \\
\midrule

OS-Atlas-4B~\cite{wu2024osatlas} & 7.1 & 0.0 & 3.7 & 3.0 & 1.4 & 2.3 & 2.0 & 0.0 & 1.5 & 9.0 & \textbf{5.5} & 7.5 & 5.1 & 3.8 & 4.4 & 5.6 & 0.0 & 3.1 & 5.0 & 1.7 & 3.7 \\
+ \emph{DiMo-GUI} & \textbf{13.6} & \textbf{1.4} & \textbf{7.7} & \textbf{9.6} & \textbf{2.8} & \textbf{6.7} & \textbf{4.1} & \textbf{4.7} & \textbf{4.2} & \textbf{30.6} & 4.5 & \textbf{19.3} & \textbf{24.3} & \textbf{15.1} & \textbf{22.2} & \textbf{7.5} & \textbf{2.2} & \textbf{5.1} & \textbf{14.6} & \textbf{4.0} & \textbf{10.6} \\
$\Delta$ & \textcolor{red}{6.5} & \textcolor{red}{1.4} & \textcolor{red}{4.0} & \textcolor{red}{6.6} & \textcolor{red}{1.4} & \textcolor{red}{4.4} & \textcolor{red}{2.1} & \textcolor{red}{4.7} & \textcolor{red}{2.7} & \textcolor{red}{21.6} & \textcolor{blue}{1.0} & \textcolor{red}{11.8} & \textcolor{red}{19.2} & \textcolor{red}{11.3} & \textcolor{red}{17.8} & \textcolor{red}{1.9} & \textcolor{red}{2.2} & \textcolor{red}{2.0} & \textcolor{red}{9.6} & \textcolor{red}{2.3} & \textcolor{red}{6.9} \\

OS-Atlas-7B~\cite{wu2024osatlas} & 33.1 & 1.4 & 17.7 & 28.8 & 2.8 & 17.9 & 12.2 & 4.7 & 10.3 & 37.5 & 7.3 & 24.4 & 33.9 & 5.7 & 27.4 & 27.1 & 4.5 & 16.8 & 28.1 & 4.0 & 18.9 \\
+ \emph{DiMo-GUI} & \textbf{66.9} & \textbf{21.4} & \textbf{44.8} & \textbf{60.6} & \textbf{21.7} & \textbf{44.3} & \textbf{50.3} & \textbf{14.1} & \textbf{41.4} & \textbf{68.1} & \textbf{21.8} & \textbf{48.0} & \textbf{80.8} & \textbf{52.8} & \textbf{74.3} & \textbf{69.2} & \textbf{28.1} & \textbf{50.5} & \textbf{65.2} & \textbf{24.5} & \textbf{49.7} \\
$\Delta$ & \textcolor{red}{33.8} & \textcolor{red}{20.0} & \textcolor{red}{27.1} & \textcolor{red}{31.8} & \textcolor{red}{18.9} & \textcolor{red}{26.4} & \textcolor{red}{38.1} & \textcolor{red}{9.4} & \textcolor{red}{31.1} & \textcolor{red}{30.6} & \textcolor{red}{14.5} & \textcolor{red}{23.6} & \textcolor{red}{46.9} & \textcolor{red}{47.1} & \textcolor{red}{46.9} & \textcolor{red}{42.1} & \textcolor{red}{23.6} & \textcolor{red}{33.7} & \textcolor{red}{37.1} & \textcolor{red}{20.5} & \textcolor{red}{30.8} \\

\midrule

UGround-7B~\cite{gou2024navigation} & 26.6 & 2.1 & 14.7 & 27.3 & 2.8 & 17.0 & 14.2 & 1.6 & 11.1 & 31.9 & 2.7 & 19.3 & 31.6 & 11.3 & 27.9 & 17.8 & 0.0 & 9.7 & 25.0 & 2.8 & 16.5 \\
+ \emph{DiMo-GUI} & \textbf{44.2} & \textbf{6.2} & \textbf{25.8} & \textbf{39.9} & \textbf{7.7} & \textbf{26.4} & \textbf{17.3} & \textbf{3.1} & \textbf{13.8} & \textbf{50.7} & \textbf{8.2} & \textbf{32.3} & \textbf{46.9} & \textbf{15.1} & \textbf{39.6} & \textbf{32.7} & \textbf{10.1} & \textbf{22.4} & \textbf{38.1} & \textbf{7.9} & \textbf{26.6} \\
$\Delta$ & \textcolor{red}{17.6} & \textcolor{red}{4.1} & \textcolor{red}{11.1} & \textcolor{red}{12.6} & \textcolor{red}{4.9} & \textcolor{red}{9.4} & \textcolor{red}{3.1} & \textcolor{red}{1.5} & \textcolor{red}{2.7} & \textcolor{red}{18.8} & \textcolor{red}{5.5} & \textcolor{red}{13.0} & \textcolor{red}{15.3} & \textcolor{red}{3.8} & \textcolor{red}{11.7} & \textcolor{red}{14.9} & \textcolor{red}{10.1} & \textcolor{red}{12.7} & \textcolor{red}{13.1} & \textcolor{red}{5.1} & \textcolor{red}{10.1} \\

UGround-V1-7B~\cite{gou2024navigation} & 51.9 & 3.4 & 28.4 & 48.0 & 9.1 & 31.7 & 20.0 & 1.6 & 15.3 & 57.6 & 16.4 & 39.8 & 61.6 & 13.2 & 50.4 & 37.4 & 7.9 & 25.0 & 45.6 & 8.4 & 31.4 \\
+ \emph{DiMo-GUI} & \textbf{57.8} & \textbf{21.4} & \textbf{40.1} & \textbf{60.1} & \textbf{18.1} & \textbf{42.5} & \textbf{45.7} & \textbf{18.8} & \textbf{39.1} & \textbf{75.7} & \textbf{28.2} & \textbf{55.1} & \textbf{79.7} & \textbf{37.7} & \textbf{70.0} & \textbf{51.4} & \textbf{30.3} & \textbf{41.8} & \textbf{61.7} & \textbf{24.3} & \textbf{47.4} \\
$\Delta$ & \textcolor{red}{5.9} & \textcolor{red}{18.0} & \textcolor{red}{11.7} & \textcolor{red}{12.1} & \textcolor{red}{9.0} & \textcolor{red}{10.8} & \textcolor{red}{25.7} & \textcolor{red}{17.2} & \textcolor{red}{23.8} & \textcolor{red}{18.1} & \textcolor{red}{11.8} & \textcolor{red}{15.3} & \textcolor{red}{18.1} & \textcolor{red}{24.5} & \textcolor{red}{19.6} & \textcolor{red}{14.0} & \textcolor{red}{22.4} & \textcolor{red}{16.8} & \textcolor{red}{16.1} & \textcolor{red}{15.9} & \textcolor{red}{16.0} \\

\bottomrule
\end{tabular}}
\vspace{-0.1in}
\caption{\textbf{Comparison of various models on ScreenSpot-Pro.} Without requiring any additional training or external data, DiMo-GUI significantly boosts the grounding performance of existing models. It nearly doubles the performance metrics of OS-ATLAS-7B and UGroundV1-7B on the ScreenSpot-Pro benchmark, with substantial improvements observed across all subsets.}
\vspace{-10pt}
\label{tab:sspro}
\end{table*}

\subsection{Modality Decoupling Strategy}
Another major challenge in GUI grounding lies in the uneven performance across different UI modalities, particularly between text-based and icon-based elements. Across multiple benchmarks, existing models consistently perform much better on text than on icons. This imbalance stems from two main issues: first, models often lack the ability to effectively recognize and understand icons, making it difficult to correctly associate them with the given instruction; second, models tend to over-rely on textual information due to their stronger language processing capabilities, often focusing on related text even when it is not the correct target. To address this issue, we propose a Modality Decoupling Strategy based on a divide-and-conquer paradigm, which explicitly separates the handling of text and icons to reduce cross-modality interference and improve grounding reliability across both modalities.

Specifically, we perform two separate grounding passes over the image: one focusing exclusively on text elements and the other on icon elements. Each pass leverages the proposed dynamic zooming mechanism to progressively refine the target location within its respective modality. After obtaining two candidate coordinate, $C_{text}$ and $C_{icon}$ from each modality, we feed them back into the model alongside the original instruction and full-resolution image. The model then evaluates both candidates and determines which coordinate is more likely to correspond to the correct target $C^*$, enabling more balanced and reliable grounding across modalities.

\begin{figure*}
    \centering
    \includegraphics[width=1.0\linewidth]{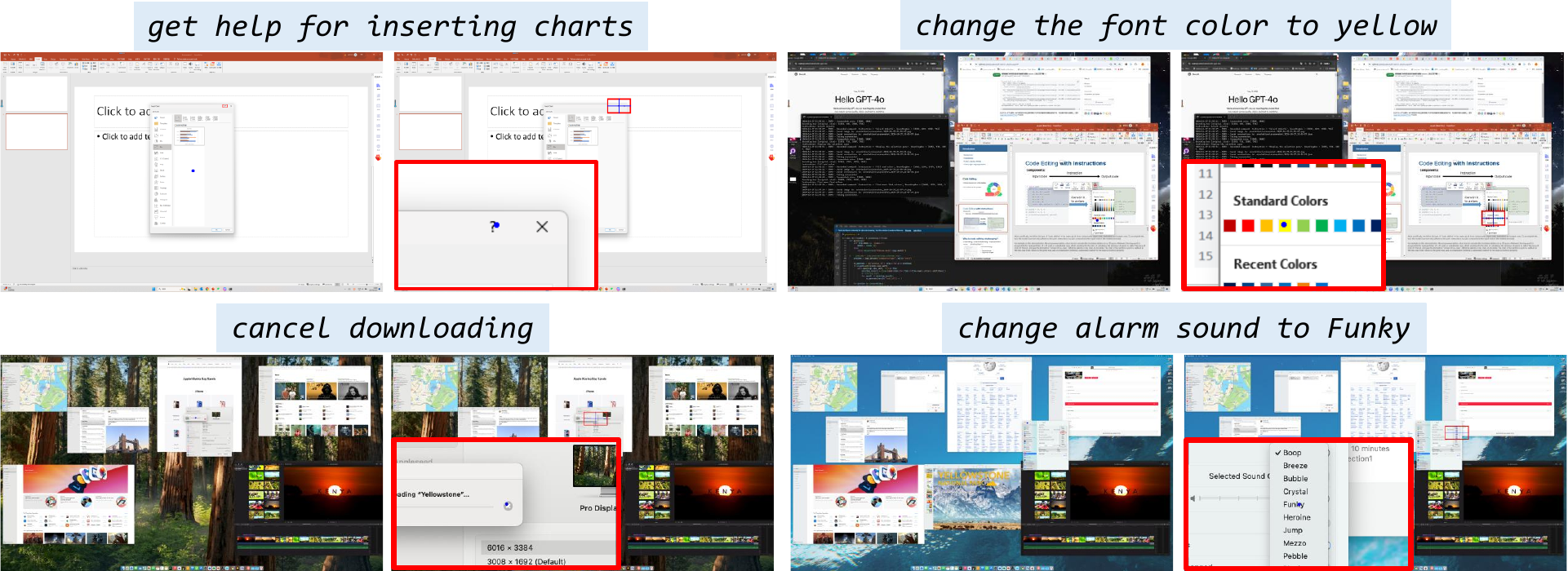}
    \caption{\textbf{Quantitative results on ScreenSpot-Pro.} On the left is the original model's prediction, where the red box represents the ground truth and the blue dot indicates the predicted coordinates. On the right is the result after integrating DiMo-GUI, where the model is able to localize more accurately according to the instruction.}
    \label{fig:sspro_results}
    \vspace{-10pt}
\end{figure*}

\section{Experiments} \label{sec:exp}
We conducted evaluations of the DiMo-GUI framework on the most recent ScreenSpot-Pro~\cite{li2024screenspot-pro} and ScreenSpot~\cite{cheng2024seeclick} benchmark datasets, and the results demonstrate its superior grounding performance compared to existing approaches.

\subsection{Experimental Setup}
\label{sec:setup}

\paragraph{Benchmarks and Models}
To thoroughly assess the grounding capabilities of DiMo-GUI, we conduct extensive experiments on two GUI grounding benchmarks: ScreenSpot~\cite{cheng2024seeclick} and ScreenSpot-Pro~\cite{li2024screenspot-pro}. ScreenSpot comprises 1,272 samples spanning mobile, desktop, and web platforms, emphasizing common interface scenarios and element types. However, due to its limited ability to represent professional software environments, ScreenSpot-Pro was introduced, featuring 23 professional applications with high-resolution interfaces and complex layouts.

On the two latest datasets mentioned above, we select the most recently reported state-of-the-art GUI agents as baseline models,~\textit{i.e.}, OS-Atlas~\cite{wu2024osatlas} and UGround-V1~\cite{gou2024navigation}. OS-Atlas is a foundational action model that leverages a multi-platform GUI grounding dataset and addresses action naming conflicts during training to enhance performance across desktop, mobile, and web platforms for GUI agent development. UGround-V1 is a universal visual grounding model for GUI agents, trained on the largest dataset of 10M GUI elements and 1.3M screenshots, utilizing web-based synthetic data and a slight adaptation of the LLaVA architecture to accurately map referring expressions to pixel-level coordinates across diverse platforms. We then apply our DiMo-GUI framework to these models to evaluate its effectiveness in enhancing the performance of GUI agent systems.

\subsection{Evaluation on Grounding Ability} 
We evaluate the effectiveness of the DiMo-GUI framework on the latest ScreenSpot-Pro dataset. As shown in Tab.~\ref{tab:sspro}, introducing the DiMo-GUI framework leads to significant performance breakthroughs for both OS-Atlas-7B and UGround-V1-7B, with OS-Atlas-7B achieving more than twice the performance of its original version. After integrating the framework, all subsets show noticeable performance improvements, demonstrating that this training-free framework delivers surprisingly strong gains in GUI grounding with minimal cost. The qualitative results further demonstrate the effectiveness of the DiMo-GUI framework. When integrated with OS-Atlas-7B and UGround-V1-7B, we observe that in the early iterations, the models often fail to return accurate coordinates—primarily due to the overwhelming contextual redundancy caused by high-resolution input. However, after several rounds of iterative zooming, the models exhibit a significantly increased likelihood of pinpointing accurate coordinates within specific regions, indicating that DiMo-GUI effectively guides the model’s attention to more relevant visual cues.

In addition, we conduct evaluations on the ScreenSpot dataset by integrating the DiMo-GUI framework into OS-Atlas-7B and UGround-V1-7B. As shown in Tab.~\ref{tab:screenspot}, both models exhibit notable performance improvements, further validating the strong generalizability of this plug-and-play framework. Despite its minimal computational cost, DiMo-GUI consistently enhances grounding performance across diverse task scenarios.

\begin{table*}[t!]
\centering
\small 
\setlength{\tabcolsep}{3pt} 
\renewcommand\arraystretch{1.1}
\begin{tabular}{cccccccc}
\hline
\multirow{2}{*}{GUI Agent MLLMs} & \multicolumn{2}{c}{Mobile} & \multicolumn{2}{c}{Desktop} & \multicolumn{2}{c}{Web} & \multirow{2}{*}{Average} \\ \cline{2-7}
 & Text & Icon/Widget & Text & Icon/Widget & Text & Icon/Widget & \\
\hline

    InternVL-2-4B~\cite{chen2024internvlscalingvisionfoundation}  & 9.2 & 4.8 & 4.6 & 4.3 & 0.9 & 0.1 & 4.3 \\
    Fuyu~\cite{bavishi2023introducing}           & 41.0 & 1.3 & 33.0 & 3.6 & 33.9 & 4.4 & 19.5 \\
    Qwen2-VL-7B~\cite{wang2024qwen2vl}    & 61.3 & 39.3 & 52.0 & 45.0 & 33.0 & 21.8 & 42.9 \\
    CogAgent~\cite{hong2024cogagent}      & 67.0 & 24.0 & 74.2 & 20.0 & 70.4 & 28.6 & 47.4 \\
    SeeClick~\cite{cheng2024seeclick}       & 78.0 & 52.0 & 72.2 & 30.0 & 55.7 & 32.5 & 53.4 \\ 
    OS-Atlas-4B~\cite{wu2024osatlas}    & 85.7 & 58.5 & 72.2 & 45.7 & 82.6 & 63.1 & 70.1 \\
    UGround-7B~\cite{gou2024navigation}     & 82.8 & 60.3 & 82.5 & 63.6 & 80.4 & 70.4 & 73.3 \\


\hline

OS-Atlas-7B~\cite{wu2024osatlas}    & 93.0 & 72.9 & 91.8 & 62.7 & \textbf{90.9} & 74.3 & 82.5 \\
+DiMo-GUI     & \textbf{96.2}~\up{3.2} & \textbf{73.5}~\up{0.6} & \textbf{96.4}~\up{4.6} & \textbf{75.1}~\up{12.4} & 89.7~\down{1.2} & \textbf{75.4}~\up{1.1} & \textbf{85.7}~\up{3.2} \\

UGround-V1-7B~\cite{gou2024navigation}   & \textbf{95.0} & 83.3 & \textbf{95.0} & 77.8 & 92.1 & 77.2 & 87.6 \\
+DiMo-GUI & 94.8~\down{0.2} & \textbf{85.3}~\up{2.0} & 94.3~\down{0.7} & \textbf{82.1}~\up{4.3} & \textbf{93.2}~\up{1.1} & \textbf{80.3}~\up{3.1} & \textbf{89.2}~\up{1.6} \\

\hline
\end{tabular}
\caption{\textbf{GUI Grounding Results of different GUI Agents on \textbf{ScreenSpot-v2}}. Even though most models already achieve high quantitative scores on this dataset, introducing DiMo-GUI still leads to noticeable performance improvements across the vast majority of subsets.}
\label{tab:screenspot}
\end{table*}


\subsection{Analysis}
In this section, we analyze the experimental results presented above to investigate the key factors that influence GUI grounding performance. By examining the strengths and weaknesses of different models across various tasks, we aim to identify the main challenges and provide insights into how future research in this field can further improve grounding accuracy and generalization. Overall, the performance of current GUI grounding models is mainly affected by two key factors: ultra-high resolution of GUI screenshots and limited visual processing ability of VLMs.

\paragraph{Ultra-high resolution of GUI screenshots}
High resolution has always been a critical issue in visual tasks. Almost all visual tasks experience a decline in performance as resolution increases, as higher resolution brings in more redundant information, making the task more challenging. GUI grounding is no exception, especially since the UI elements that need to be localized are often small. As shown in Figure 1, performance in GUI grounding significantly drops as the resolution increases. An intuitive solution to this issue is zooming in, which is the dynamic zooming approach proposed in this paper. However, it can be observed that as the resolution of the screenshots increases, the probability of the model making errors in the first iteration also increases, which inevitably leads to failure in subsequent operations. On the contrary, blindly enlarging the image can also introduce negative effects—for instance, excessive magnification may lead to a loss of global information. Determining the appropriate degree of magnification plays a crucial role in the task of GUI grounding, making a dynamic zooming strategy essential.

\begin{figure*}
    \centering
    \includegraphics[width=1.0\linewidth]{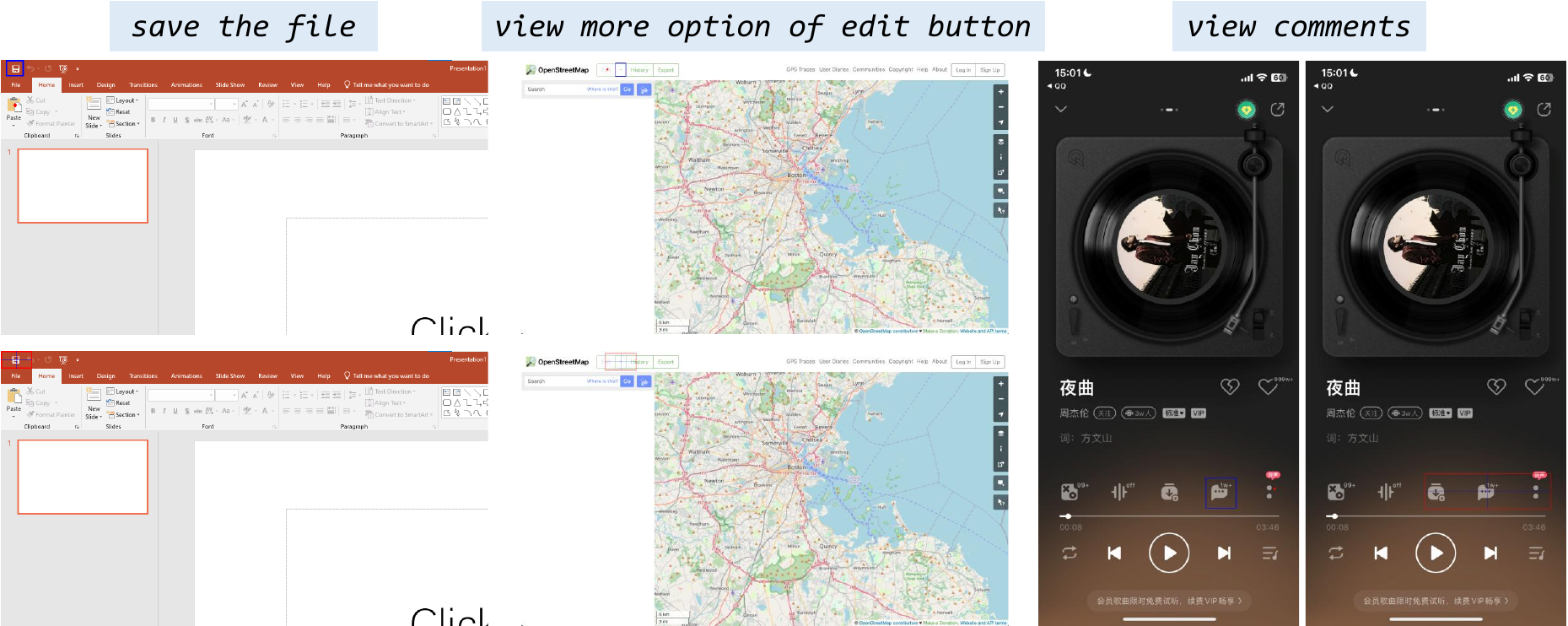}
    \caption{\textbf{Quantitative results on ScreenSpot-v2.} On the Screenspot benchmark, which features relatively low resolution and simple scenes, DiMo-GUI also enhances the model's localization capabilities.}
    \vspace{-10pt}
    \label{fig:ss_results}
\end{figure*}

\paragraph{Limited visual processing ability of VLMs}
Another reason for the poor performance of GUI grounding is the weak ability of grounding models to process visual information. Most current GUI agents and grounding models are based on existing multimodal large models, and a common issue with MLLMs is that their ability to process visual information is weaker than their ability to handle text. This causes the models to be more inclined to trust textual information, a phenomenon known as hallucinations in MLLMs. Since locating, recognizing, and understanding icons is much more difficult than processing text, GUI agents tend to rely more on textual information during the grounding process. The direct consequence is that if a screenshot contains text related to the instruction, or even the same text, GUI agents will almost completely abandon the search for icons and instead use the text as the answer, even though it may not be helpful. The modality decoupling approach we propose effectively addresses this issue by allowing the model to better consider both text and icon modalities, which helps mitigate the drawbacks of the model's weaker ability to process visual information.

\begin{figure}
    \centering
    \includegraphics[width=0.9\linewidth]{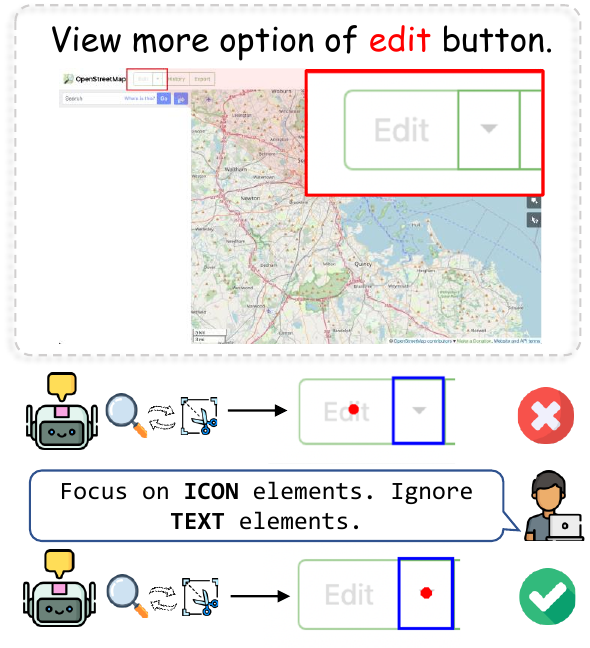}
    \caption{\textbf{Case Study.} GUI agents often mistake instruction-related text in the image as targets. Using a divide-and-conquer approach with explicit modality helps the agent locate the target accurately.}
    \vspace{-5pt}
    \label{fig:case}
\end{figure}

As illustrated in the specific example in Fig.~\ref{fig:case}, when the user instruction includes the word “edit,” the agent tends to focus on elements related to editing during the search process. In this case, there happens to be a text element labeled “Edit” in the target region, which conveys a clearer semantic meaning compared to the adjacent icon. Consequently, the agent model is more likely to rely on this text element, as it is not only easier to recognize and understand but also highly relevant to the instruction. However, this text element does not actually fulfill the intended function of the instruction. Its seemingly clear semantics, in this context, become a source of distraction. When we modify the prompt to explicitly direct the agent to focus only on icon elements while ignoring text elements, the model  DiMo-GUIs selecting the “Edit” text and instead searches for the appropriate icon. Interestingly, the “Edit” text then serves as valuable contextual information that aids the model in locating the target icon—transforming from a source of distraction into a helpful cue.

\begin{table}[t]
\centering
\caption{Ablation on the number of iterative zooming steps. Performance improves with more iterations, but plateaus after 3 steps.}
\vspace{-5pt}
\label{tab:ablation_zooming}
\resizebox{\columnwidth}{!}{
\begin{tabular}{l | c c c c c c}
\toprule
max\_iter & 0 & 1 & 2 & 3 & 4 & 5 \\
\midrule
acc (\%) & 18.4 & 18.7 & 40.2 & 46.7 & 48.8 & \textbf{48.9} \\
\bottomrule
\end{tabular}
}
\vspace{-10pt}
\end{table}



\begin{table}[t]
\centering
\small
\caption{Ablation on Dynamic Grounding and Modalities Dividing.}
\vspace{-5pt}
\label{tab:ablation_modality}
\resizebox{0.5\columnwidth}{!}{%
\begin{tabular}{l|c}
\toprule
Method & OS-Atlas-7B \\
\midrule
vanilla & 18.4 \\
w DG & 45.7 \\
w MD & 26.1 \\
\textbf{w DiMo-GUI} & \textbf{49.7} \\
\bottomrule
\end{tabular}%
}
\vspace{-10pt}
\end{table}

\subsection{Ablation Study}

\paragraph{Ablation on Dynamic Zooming}
To validate the effectiveness of the proposed dynamic zooming strategy, we compare DiMo-GUI with two baselines: (1) a no-zooming baseline where the model directly predicts coordinates from the original screenshot without any refinement, and (2) a single-pass static zooming variant that only zooms into the region of interest once based on the initial prediction. As shown in Tab.~\ref{tab:ablation_zooming}, the grounding performance first improves and then declines with the monotonic increase in iterations.This aligns with intuition: in early stages, more zoom-in operations help the model focus on target regions by filtering out irrelevant details. However, excessive zooming can remove important context, hindering accurate grounding. Our proposed dynamic iterative zooming approach significantly improves grounding accuracy over both baselines, which demonstrates the importance of progressively refining the region of interest.

\paragraph{Ablation on Modality Decoupling}
We also investigate the impact of modality decoupling by comparing the full DiMo-GUI framework with a variant that treats all UI elements uniformly without distinguishing between text and icon modalities. The results in Tab.~\ref{tab:ablation_modality} show that modality-aware processing leads to consistent performance gains. This confirms our hypothesis that different modalities benefit from specialized zooming strategies, and that decoupling helps reduce visual ambiguity, particularly in scenarios where icons are harder to interpret than text.

\section{Conclusion}
DiMo-GUI is a training-free, plug-and-play framework designed specifically for the GUI grounding task. It incorporates two key components: dynamic zooming and modality decoupling, which effectively address the challenges of handling high-resolution screenshots and the limited visual understanding capability of existing GUI agents. By progressively refining the focus region and treating text and icon modalities separately, DiMo-GUI significantly boosts grounding performance across various benchmarks and models, offering substantial improvements with minimal computational overhead.

\section{Limitations}
Currently our model employs a progressive expansion strategy without any error correction or backtracking mechanisms. This can lead to early-stage mistakes that propagate and become irrecoverable. In future work, we plan to incorporate backtracking mechanisms using structures such as trees or graphs, aiming to further improve the accuracy.

\section{Acknowledgments}
The work is partially supported by the NSF of the United States Grant CRII 2451683, an NVIDIA Academic Grants Program, University of California at Merced, and a UC Merced Faculty Research Award.
The views and conclusions are those of the authors and should not reflect the official policy or position of the U.S. Government.

\bibliography{acl2020}

\begin{thebibliography}{38}
\expandafter\ifx\csname natexlab\endcsname\relax\def\natexlab#1{#1}\fi

\bibitem[{Anderson et~al.(2018)Anderson, Wu, Teney, Bruce, Johnson, S{\"u}nderhauf, Reid, Gould, and Van Den~Hengel}]{anderson2018vision}
Peter Anderson, Qi~Wu, Damien Teney, Jake Bruce, Mark Johnson, Niko S{\"u}nderhauf, Ian Reid, Stephen Gould, and Anton Van Den~Hengel. 2018.
\newblock Vision-and-language navigation: Interpreting visually-grounded navigation instructions in real environments.
\newblock \emph{Proceedings of the IEEE/CVF Conference on Computer Vision and Pattern Recognition (CVPR)}, pages 3674--3683.

\bibitem[{Bai et~al.(2023)Bai, Bai, Yang, Wang, Tan, Wang, Lin, Zhou, and Zhou}]{bai2023qwenvl}
Jinze Bai, Shuai Bai, Shusheng Yang, Shijie Wang, Sinan Tan, Peng Wang, Junyang Lin, Chang Zhou, and Jingren Zhou. 2023.
\newblock \href {http://arxiv.org/abs/2308.12966} {Qwen-vl: A versatile vision-language model for understanding, localization, text reading, and beyond}.

\bibitem[{Bavishi et~al.(2023)Bavishi, Elsen, Hawthorne, Nye, Odena, Somani, and Ta\c{s}{\i}rlar}]{bavishi2023introducing}
Rohan Bavishi, Erich Elsen, Curtis Hawthorne, Maxwell Nye, Augustus Odena, Arushi Somani, and Sagnak Ta\c{s}{\i}rlar. 2023.
\newblock Introducing our multimodal models.

\bibitem[{Chen et~al.(2024)Chen, Wu, Wang, Su, Chen, Xing, Zhong, Zhang, Zhu, Lu, Li, Luo, Lu, Qiao, and Dai}]{chen2024internvlscalingvisionfoundation}
Zhe Chen, Jiannan Wu, Wenhai Wang, Weijie Su, Guo Chen, Sen Xing, Muyan Zhong, Qinglong Zhang, Xizhou Zhu, Lewei Lu, Bin Li, Ping Luo, Tong Lu, Yu~Qiao, and Jifeng Dai. 2024.
\newblock \href {http://arxiv.org/abs/2312.14238} {{InternVL}: Scaling up vision foundation models and aligning for generic visual-linguistic tasks}.

\bibitem[{Cheng et~al.(2024)Cheng, Sun, Chu, Xu, Li, Zhang, and Wu}]{cheng2024seeclick}
Kanzhi Cheng, Qiushi Sun, Yougang Chu, Fangzhi Xu, Yantao Li, Jianbing Zhang, and Zhiyong Wu. 2024.
\newblock Seeclick: Harnessing gui grounding for advanced visual gui agents.
\newblock \emph{Annual Meeting of the Association for Computational Linguistics (ACL)}.

\bibitem[{Du et~al.(2020)Du, Li, Guo, Yin, Liu, Zhou, Bai, Yu, Yang, Dang, and Wang}]{du2020ppocr}
Yuning Du, Chenxia Li, Ruoyu Guo, Xiaoting Yin, Weiwei Liu, Jun Zhou, Yifan Bai, Zilin Yu, Yehua Yang, Qingqing Dang, and Haoshuang Wang. 2020.
\newblock \href {http://arxiv.org/abs/2009.09941} {{PP-OCR}: A practical ultra lightweight ocr system}.

\bibitem[{Ge et~al.(2025)Ge, Wang, Yang, and Cai}]{ge2025mrfd}
Haonan Ge, Yiwei Wang, Ming-Hsuan Yang, and Yujun Cai. 2025.
\newblock Mrfd: Multi-region fusion decoding with self-consistency for mitigating hallucinations in lvlms.
\newblock \emph{arXiv preprint arXiv:2508.10264}.

\bibitem[{Gou et~al.(2024)Gou, Wang, Zheng, Xie, Chang, Shu, Sun, and Su}]{gou2024navigation}
Boyu Gou, Ruohan Wang, Boyuan Zheng, Yanan Xie, Cheng Chang, Yiheng Shu, Huan Sun, and Yu~Su. 2024.
\newblock Navigating the digital world as humans do: Universal visual grounding for gui agents.
\newblock \emph{International Conference on Learning Representations (ICLR)}.

\bibitem[{Hong et~al.(2024)Hong, Wang, Lv, Xu, Yu, Ji, Wang, Wang, Zhang, Li, Xu, Dong, Ding, and Tang}]{hong2024cogagent}
Wenyi Hong, Weihan Wang, Qingsong Lv, Jiazheng Xu, Wenmeng Yu, Junhui Ji, Yan Wang, Zihan Wang, Yuxuan Zhang, Juanzi Li, Bin Xu, Yuxiao Dong, Ming Ding, and Jie Tang. 2024.
\newblock \href {http://arxiv.org/abs/2312.08914} {{CogAgent}: A visual language model for gui agents}.

\bibitem[{Hosseini et~al.(2024)Hosseini, Yuan, Malkin, Courville, Sordoni, and Agarwal}]{hosseini2024vst}
Arian Hosseini, Xingdi Yuan, Nikolay Malkin, Aaron Courville, Alessandro Sordoni, and Rishabh Agarwal. 2024.
\newblock V-star: Training verifiers for self-taught reasoners.
\newblock \emph{arXiv preprint arXiv:2402.06457}.

\bibitem[{Hu et~al.(2024)Hu, Ouyang, Gao, and Shou}]{hu2024claude}
Siyuan Hu, Mingyu Ouyang, Difei Gao, and Mike~Zheng Shou. 2024.
\newblock \href {http://arxiv.org/abs/2411.10323} {{The Dawn of GUI Agent}: A preliminary case study with claude 3.5 computer use}.

\bibitem[{Lee et~al.(2025{\natexlab{a}})Lee, Kim, Kim, Tack, Jo, Lee, Park, In, Shin, and Yoo}]{lee2025reguide}
Hyunseok Lee, Jeonghoon Kim, Beomjun Kim, Jihoon Tack, Chansong Jo, Jaehong Lee, Cheonbok Park, Sookyo In, Jinwoo Shin, and Kang~Min Yoo. 2025{\natexlab{a}}.
\newblock {ReGUIDE}: Data efficient gui grounding via spatial reasoning and search.
\newblock \emph{arXiv preprint arXiv:2505.15259}.

\bibitem[{Lee et~al.(2025{\natexlab{b}})Lee, Oh, Kim, Shin, and Tack}]{lee2025revise}
Hyunseok Lee, Seunghyuk Oh, Jaehyung Kim, Jinwoo Shin, and Jihoon Tack. 2025{\natexlab{b}}.
\newblock Revise: Learning to refine at test-time via intrinsic self-verification.
\newblock \emph{arXiv preprint arXiv:2502.14565}.

\bibitem[{Li et~al.(2025)Li, Meng, Lin, Luo, Tian, Ma, Huang, and Chua}]{li2024screenspot-pro}
Kaixin Li, Ziyang Meng, Hongzhan Lin, Ziyang Luo, Yuchen Tian, Jing Ma, Zhiyong Huang, and Tat-Seng Chua. 2025.
\newblock {ScreenSpot-Pro}: Gui grounding for professional high-resolution computer use.

\bibitem[{Lin et~al.(2024)Lin, Li, Gao, Yang, Wu, Bai, Lei, Wang, and Shou}]{lin2024showui}
Kevin~Qinghong Lin, Linjie Li, Difei Gao, Zhengyuan Yang, Shiwei Wu, Zechen Bai, Weixian Lei, Lijuan Wang, and Mike~Zheng Shou. 2024.
\newblock \href {http://arxiv.org/abs/2411.17465} {{ShowUI}: One vision-language-action model for gui visual agent}.

\bibitem[{Liu et~al.(2024{\natexlab{a}})Liu, Zeng, Ren, Li, Zhang, Yang, Jiang, Li, Yang, Su, Zhu, and Zhang}]{liu2024groundingdino}
Shilong Liu, Zhaoyang Zeng, Tianhe Ren, Feng Li, Hao Zhang, Jie Yang, Qing Jiang, Chunyuan Li, Jianwei Yang, Hang Su, Jun Zhu, and Lei Zhang. 2024{\natexlab{a}}.
\newblock \href {http://arxiv.org/abs/2303.05499} {{Grounding DINO}: Marrying dino with grounded pre-training for open-set object detection}.

\bibitem[{Liu et~al.(2024{\natexlab{b}})Liu, Zhang, Gu, Iong, Xu, Song, Zhang, Lai, Liu, Zhao et~al.}]{liu2024visualagentbench}
Xiao Liu, Tianjie Zhang, Yu~Gu, Iat~Long Iong, Yifan Xu, Xixuan Song, Shudan Zhang, Hanyu Lai, Xinyi Liu, Hanlin Zhao, et~al. 2024{\natexlab{b}}.
\newblock {VisualAgentBench}: Towards large multimodal models as visual foundation agents.
\newblock \emph{arXiv preprint arXiv:2408.06327}.

\bibitem[{Lu et~al.(2024)Lu, Yang, Shen, and Awadallah}]{lu2024omniparser}
Yadong Lu, Jianwei Yang, Yelong Shen, and Ahmed Awadallah. 2024.
\newblock Omniparser for pure vision based gui agent.
\newblock \emph{arXiv preprint arXiv:2408.00203}.

\bibitem[{Luo et~al.(2025)Luo, Logeswaran, Johnson, and Lee}]{luo2025visual}
Tiange Luo, Lajanugen Logeswaran, Justin Johnson, and Honglak Lee. 2025.
\newblock Visual test-time scaling for gui agent grounding.
\newblock \emph{arXiv preprint arXiv:2505.00684}.

\bibitem[{Nakano et~al.(2022)Nakano, Hilton, Balaji, Wu, Ouyang, Kim, Hesse, Jain, Kosaraju, Saunders, Jiang, Cobbe, Eloundou, Krueger, Button, Knight, Chess, and Schulman}]{nakano2022webgpt}
Reiichiro Nakano, Jacob Hilton, Suchir Balaji, Jeff Wu, Long Ouyang, Christina Kim, Christopher Hesse, Shantanu Jain, Vineet Kosaraju, William Saunders, Xu~Jiang, Karl Cobbe, Tyna Eloundou, Gretchen Krueger, Kevin Button, Matthew Knight, Benjamin Chess, and John Schulman. 2022.
\newblock \href {http://arxiv.org/abs/2112.09332} {{WebGPT}: Browser-assisted question-answering with human feedback}.

\bibitem[{Nguyen(2024)}]{nguyen2024improved}
Anthony Nguyen. 2024.
\newblock Improved gui grounding via iterative narrowing.
\newblock \emph{arXiv preprint arXiv:2411.13591}.

\bibitem[{OpenAI(2023)}]{openai2023gpt4}
OpenAI. 2023.
\newblock \href {http://arxiv.org/abs/2303.08774} {Gpt-4 technical report}.

\bibitem[{Qin et~al.(2025)Qin, Ye, Fang, Wang, Liang, Tian, Zhang, Li, Li, Huang et~al.}]{qin2025uitars}
Yujia Qin, Yining Ye, Junjie Fang, Haoming Wang, Shihao Liang, Shizuo Tian, Junda Zhang, Jiahao Li, Yunxin Li, Shijue Huang, et~al. 2025.
\newblock {UI-TARS}: Pioneering automated gui interaction with native agents.
\newblock \emph{arXiv preprint arXiv:2501.12326}.

\bibitem[{Snell et~al.(2024)Snell, Lee, Xu, and Kumar}]{snell2024scaling}
Charlie Snell, Jaehoon Lee, Kelvin Xu, and Aviral Kumar. 2024.
\newblock Scaling llm test-time compute optimally can be more effective than scaling model parameters.
\newblock \emph{arXiv preprint arXiv:2408.03314}.

\bibitem[{Tang et~al.(2025)Tang, Shen, Zhang, Chen, Hou, Zhang, Zhang, Song, Lu, and Zhuang}]{tang2025think}
Fei Tang, Yongliang Shen, Hang Zhang, Siqi Chen, Guiyang Hou, Wenqi Zhang, Wenqiao Zhang, Kaitao Song, Weiming Lu, and Yueting Zhuang. 2025.
\newblock {Think Twice, Click Once}: Enhancing gui grounding via fast and slow systems.
\newblock \emph{arXiv preprint arXiv:2503.06470}.

\bibitem[{Tao et~al.(2025)Tao, Wang, Cai, Yang, and Tang}]{tao2025understanding}
Xingjian Tao, Yiwei Wang, Yujun Cai, Zhicheng Yang, and Jing Tang. 2025.
\newblock Understanding gui agent localization biases through logit sharpness.
\newblock \emph{arXiv preprint arXiv:2506.15425}.

\bibitem[{Wang et~al.(2024{\natexlab{a}})Wang, Xu, Ye, Yan, Shen, Zhang, Huang, and Sang}]{wang2024mobileagent}
Junyang Wang, Haiyang Xu, Jiabo Ye, Ming Yan, Weizhou Shen, Ji~Zhang, Fei Huang, and Jitao Sang. 2024{\natexlab{a}}.
\newblock \href {http://arxiv.org/abs/2401.16158} {{Mobile-Agent}: Autonomous multi-modal mobile device agent with visual perception}.

\bibitem[{Wang et~al.(2024{\natexlab{b}})Wang, Bai, Tan, Wang, Fan, Bai, Chen, Liu, Wang, Ge, Fan, Dang, Du, Ren, Men, Liu, Zhou, Zhou, and Lin}]{wang2024qwen2vl}
Peng Wang, Shuai Bai, Sinan Tan, Shijie Wang, Zhihao Fan, Jinze Bai, Keqin Chen, Xuejing Liu, Jialin Wang, Wenbin Ge, Yang Fan, Kai Dang, Mengfei Du, Xuancheng Ren, Rui Men, Dayiheng Liu, Chang Zhou, Jingren Zhou, and Junyang Lin. 2024{\natexlab{b}}.
\newblock \href {http://arxiv.org/abs/2409.12191} {Qwen2-vl: Enhancing vision-language model's perception of the world at any resolution}.

\bibitem[{Wu and Xie(2024)}]{wu2024vstar}
Penghao Wu and Saining Xie. 2024.
\newblock V*: Guided visual search as a core mechanism in multimodal llms.
\newblock \emph{Proceedings of the IEEE/CVF Conference on Computer Vision and Pattern Recognition (CVPR)}, pages 13084--13094.

\bibitem[{Wu et~al.(2024)Wu, Wu, Xu, Wang, Sun, Jia, Cheng, Ding, Chen, Liang et~al.}]{wu2024osatlas}
Zhiyong Wu, Zhenyu Wu, Fangzhi Xu, Yian Wang, Qiushi Sun, Chengyou Jia, Kanzhi Cheng, Zichen Ding, Liheng Chen, Paul~Pu Liang, et~al. 2024.
\newblock Os-atlas: A foundation action model for generalist gui agents.
\newblock \emph{International Conference on Learning Representations (ICLR)}.

\bibitem[{Xia and Luo(2025)}]{xia2025gui}
Xiaobo Xia and Run Luo. 2025.
\newblock {GUI-R1}: A generalist r1-style vision-language action model for gui agents.
\newblock \emph{arXiv preprint arXiv:2504.10458}.

\bibitem[{Yang et~al.(2024{\natexlab{a}})Yang, Liu, Chaudhary, Fakoor, Chaudhari, Karypis, and Rangwala}]{yang2024agentoccam}
Ke~Yang, Yao Liu, Sapana Chaudhary, Rasool Fakoor, Pratik Chaudhari, George Karypis, and Huzefa Rangwala. 2024{\natexlab{a}}.
\newblock Agentoccam: A simple yet strong baseline for llm-based web agents.
\newblock \emph{arXiv preprint arXiv:2410.13825}.

\bibitem[{Yang et~al.(2024{\natexlab{b}})Yang, Wang, Li, Luo, Chen, Huang, and Li}]{yang2024ariaui}
Yuhao Yang, Yue Wang, Dongxu Li, Ziyang Luo, Bei Chen, Chao Huang, and Junnan Li. 2024{\natexlab{b}}.
\newblock \href {http://arxiv.org/abs/2412.16256} {{Aria-UI}: Visual grounding for gui instructions}.

\bibitem[{Yuan et~al.(2025)Yuan, Zhang, Li, Cai, Yao, Chen, Wang, Hou, Chen, Jiang et~al.}]{yuan2025enhancing}
Xinbin Yuan, Jian Zhang, Kaixin Li, Zhuoxuan Cai, Lujian Yao, Jie Chen, Enguang Wang, Qibin Hou, Jinwei Chen, Peng-Tao Jiang, et~al. 2025.
\newblock Enhancing visual grounding for gui agents via self-evolutionary reinforcement learning.
\newblock \emph{arXiv preprint arXiv:2505.12370}.

\bibitem[{Zhang et~al.(2024)Zhang, Li, He, Zhang, Qiao, Qin, Ma, Kang, Lin, Rajmohan, Zhang, and Zhang}]{zhang2024ufo}
Chaoyun Zhang, Liqun Li, Shilin He, Xu~Zhang, Bo~Qiao, Si~Qin, Minghua Ma, Yu~Kang, Qingwei Lin, Saravan Rajmohan, Dongmei Zhang, and Qi~Zhang. 2024.
\newblock \href {http://arxiv.org/abs/2402.07939} {{UFO}: A ui-focused agent for windows os interaction}.

\bibitem[{Zhang et~al.(2023)Zhang, Yang, Liu, Han, Chen, Huang, Fu, and Yu}]{zhang2023appagent}
Chi Zhang, Zhao Yang, Jiaxuan Liu, Yucheng Han, Xin Chen, Zebiao Huang, Bin Fu, and Gang Yu. 2023.
\newblock \href {http://arxiv.org/abs/2312.13771} {{AppAgent}: Multimodal agents as smartphone users}.

\bibitem[{Zhang and Zhang(2023)}]{zhang2023you}
Zhuosheng Zhang and Aston Zhang. 2023.
\newblock You only look at screens: Multimodal chain-of-action agents.
\newblock \emph{Annual Meeting of the Association for Computational Linguistics (ACL)}.

\bibitem[{Zhou et~al.(2025)Zhou, Dai, Wang, Zhou, Jia et~al.}]{zhou2025gui}
Yuqi Zhou, Sunhao Dai, Shuai Wang, Kaiwen Zhou, Qinqlin Jia, et~al. 2025.
\newblock {GUI-G1}: Understanding r1-zero-like training for visual grounding in gui agents.
\newblock \emph{arXiv preprint arXiv:2505.15810}.

\end{thebibliography}

\afterpage{\blankpage}

\newpage
\appendix



\end{document}